\documentclass[letterpaper]{article} % DO NOT CHANGE THIS

\usepackage{aaai2027}  % DO NOT CHANGE THIS

\usepackage[utf8]{inputenc}
\usepackage[hyphens]{url}  % DO NOT CHANGE THIS
\usepackage{graphicx} % DO NOT CHANGE THIS
\usepackage{natbib}  % DO NOT CHANGE THIS AND DO NOT ADD ANY OPTIONS TO IT
\usepackage{caption} % DO NOT CHANGE THIS AND DO NOT ADD ANY OPTIONS TO IT
\usepackage{amsmath}
\usepackage{amssymb}
\usepackage{mathtools}
\usepackage{amsthm}
\usepackage{booktabs}
\usepackage{multirow}
\usepackage{makecell}
\usepackage{pifont}
\usepackage{subcaption}
\usepackage{algorithm}
\usepackage{algorithmic}
\newcommand{\cmark}{\ding{51}}
\newcommand{\xmark}{\ding{55}}

\theoremstyle{plain}

\theoremstyle{definition}

\theoremstyle{remark}

\title{Separate First, Fuse Later: Mitigating Cross-Modal Interference\\in Audio-Visual LLMs Reasoning with Modality-Specific Chain-of-Thought}

\author{
    Xuanchen Li\textsuperscript{\rm 1},
    Yuheng Lu\textsuperscript{\rm 1},
    Chenrui Cui\textsuperscript{\rm 1},
    Tianrui Wang\textsuperscript{\rm 1},
    Zikang Huang\textsuperscript{\rm 1},
    Yu Jiang\textsuperscript{\rm 1},\\
    Long Zhou\textsuperscript{\rm 4},
    Longbiao Wang\textsuperscript{\rm 1,\rm 2}\corresponding,
    Jianwu Dang\textsuperscript{\rm 1,\rm 3}
}
\affiliations{
    \textsuperscript{\rm 1}Tianjin Key Laboratory of Cognitive Computing and Application, Tianjin University, Tianjin, China\\
    \textsuperscript{\rm 2}Huiyan Technology Company, Ltd., Tianjin, China\\
    \textsuperscript{\rm 3}Chinese Academy of Sciences, Guangdong, China\\
    \textsuperscript{\rm 4}Independent Researcher\\
    longbiao\_wang@tju.edu.cn
}

\begin{document}

\maketitle

\begin{abstract}
Audio and vision provide complementary evidence for audio-visual question answering, yet current audio-visual large language models may suffer from cross-modal interference: information from one modality misguides the interpretation of another, thereby inducing hallucinations.
We attribute this issue to uncontrolled cross-modal interactions during intermediate reasoning.
To mitigate this, we propose Separate First, Fuse Later (SFFL), an audio-visual reasoning framework designed to reduce cross-modal interference. SFFL enforces modality-specific chain-of-thought reasoning, producing separate audio and visual reasoning traces and integrating evidence for answering.
We construct modality-preference labels via a data pipeline under different modality input settings. We use these labels as an auxiliary reward in reinforcement learning to encourage an instance-dependent preference for modality cues when answering.
We further introduce a modality-specific reasoning mechanism that preserves modality isolation during the separated reasoning stage while enabling full access to cross-modal information at the evidence fusion stage.
Experiments demonstrate consistent improvements in both accuracy and robustness, yielding an average relative gain of 5.16\% on general AVQA benchmarks and 11.17\% on a cross-modal hallucination benchmark.

\end{abstract}
% model producing separate audio and visual rationales and summarizing evidence for answering.
% SFFL enforces model explicitly separates audio and video reasoning and summarizes each modality’s independent evidence.
%  at the final decision stage.
% The model is trained end-to-end using reinforcement learning to jointly encourage structured reasoning and improve accuracy.
% we construct modality-preference labels by evaluating answer correctness and cross-setting agreement under audio-only, visual-only, and audio-visual inputs, and use them as training signals for reinforcement learning.
% AVQA requires models to jointly reason over visual and auditory inputs, identify question-relevant evidence, and capture cross-modal relationships to answer diverse questions.
% However, naively providing audio and video jointly does not always improve performance~\cite{mo1,mo2,avh}. In practice, we observe a counterintuitive failure mode of AVLMs: When a single modality is sufficient to solve the question, introducing an additional modality can paradoxically reduce accuracy and further trigger cross-modal hallucinations. (i.e., audio-visual LLMs may hear imaginary sounds from visual cues or perceive fake visual events from audio cues).
% We attribute this phenomenon to two key issues: (1) Asymmetric information density across modalities, which biases attention toward the denser modality~\cite{as1,as2,as3}, and (2) Uncontrolled cross-modal information flow in autoregressive reasoning lets noise or shortcut cues in one modality interfere with the other-modality reasoning~\cite{inter1,inter2}.
\section{Introduction}
\label{sec:introduction} 
% Fig\ref{}.
\begin{figure}[t]
  \centering
  \includegraphics[width=1.0\columnwidth]{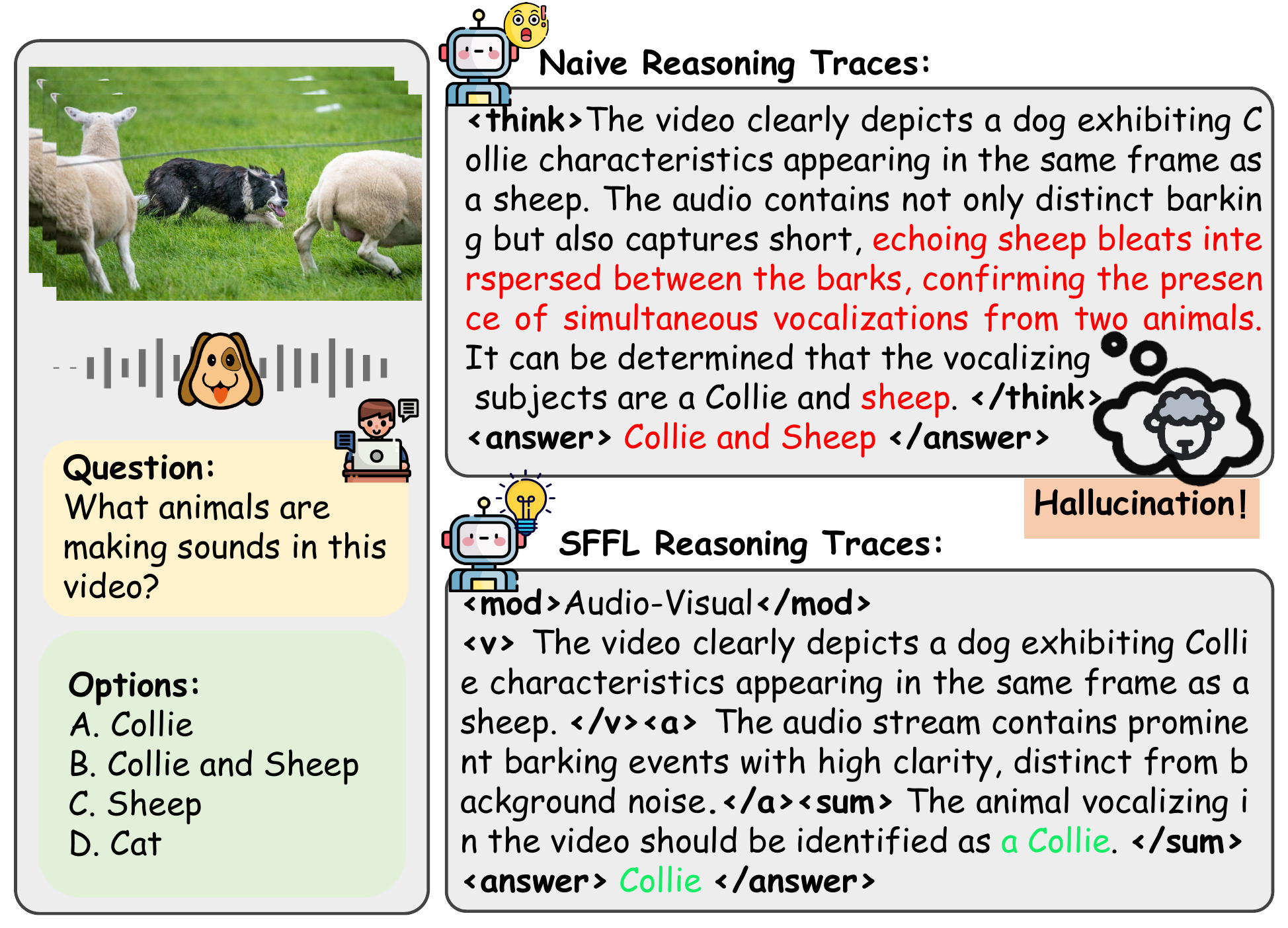}
  \caption{This figure shows cross-modal interference and how SFFL mitigates it: although a collie and sheep appear in the frame, the audio contains only barking. Naive joint reasoning wrongly attributes the sound to both animals, while SFFL separates audio/visual reasoning and fuses evidence only at the end, correctly identifying only the collie.}
  \label{fig:intro}
\end{figure}

Humans understand the world by jointly interpreting visual and auditory signals~\cite{merge, merge2}.  
Visual information conveys objects, agents, and spatial relationships, while audio provides complementary cues such as sound sources and event characteristics~\cite{a}.
Motivated by this complementary perception, audio-visual question answering (AVQA) has attracted growing attention as a task that requires models to accurately identify question-relevant evidence from audio and video inputs and reason across modalities~\cite{avqam1,avqam2,avqam3}.
With the progress of large language models in multimodal learning, audio-visual large language models (AVLMs)~\cite{gpt4o,vl2,vs2,qwen3o}  have demonstrated strong capabilities in integrating visual and auditory cues for audio-visual understanding.
Most existing approaches encode visual and audio inputs separately and then feed tokens from both modalities into a single LLM for joint reasoning~\cite{vl2,vs2,qwen3o}.
These methods achieve competitive performance on standard AVQA benchmarks~\cite{avqa,musicavqa,valor2}.

However, jointly providing audio and video does not always lead to better performance and has been widely observed to introduce cross-modal interference and hallucinations~\cite{mo1,mo2,avh}. 
In particular, information from one modality can improperly bias the interpretation of the other, causing the model to infer content that is not grounded in the queried modality.
For example, the presence of a sheep in the visual scene may lead the model to attribute a bleating sound to it, even when the audio provides no evidence of such a sound (see Fig~\ref{fig:intro}).
We attribute this issue primarily to two factors. First, most existing AVLMs concatenate audio and visual features and perform joint reasoning over the combined sequence. This lack of explicit control over modality-specific cue usage allows irrelevant signals from one modality to interfere with the other’s reasoning~\cite{inter1,inter2,as3}.
Second, prior studies show that current AVLMs exhibit a visual-dominant bias in multimodal reasoning, often relying primarily on visual cues while underutilizing auditory evidence~\cite{as1,as2,bias1}.

To mitigate these challenges, we propose Separate First, Fuse Later (SFFL), a structured reasoning paradigm that separates modality-specific processing from cross-modal integration.
Unlike conventional methods that rely on uncontrolled joint reasoning, SFFL first performs independent reasoning on audio and visual streams. It then fuses these evidence traces only in the final generation stage, which helps mitigate cross-modal interference during earlier reasoning steps.

Our main contributions are summarized as follows:

\begin{itemize}
  \item We propose the SFFL framework, incorporating a modality-specific reasoning mechanism that enforces isolation during the chain-of-thought stage. This design prevents premature information mixing while enabling full access to cross-modal information at the evidence fusion stage.
  \item We design a modality-preference data pipeline. By analyzing answer correctness and consistency across different modality input settings. We use these labels as an auxiliary reward in reinforcement learning to encourage an instance-dependent preference for modality cues when answering.
  \item Extensive experiments show that SFFL achieves the best aggregate performance among the compared methods with complete results, with average relative gains of 5.16\% on AVQA and 11.17\% on cross-modality hallucination benchmarks. An anonymized demo is available at https://anon7f3c2a.github.io/.
\end{itemize}

% Specifically, we introduce Perferred Evidence Modality (PEM) prediction to . To obtain supervision for PEM, we construct a self-distillation labeling pipeline on AVQA: for each question, we perform multiple sampled reasoning runs under three input settings (A/V/AV), and determine whether each setting is solvable based on answer correctness and reasoning consistency. We then assign the PEM label as one of Audio/Visual/Audio-Visual accordingly.
% We further propose Separate AV Thinking (SAVT), which decomposes perception and reasoning into independent audio and video segments and integrates evidence at the end. To address the implicit “fully-visible history” information-sharing pathway in conventional unified autoregressive fusion, we introduce MAAM, which imposes asymmetric visibility constraints across reasoning segments during autoregressive generation, explicitly preventing cross-modal information leakage and cross-segment conclusion leakage.
% Finally, since these behavioral constraints are difficult to reliably acquire with supervised learning alone, we apply GRPO for optimization, using rewards that jointly enforce modality consistency/format compliance and final answer correctness. Overall, our framework directly targets the two failure sources above and substantially improves robust audio-visual reasoning under joint audio-video input.

\section{Related Work}
\label{sec:related_work} 
\subsection{Audio-Visual Large Language Models} 
\label{sec:rel_avlm} 
Recent advancements in Audio-Visual Large Language Models (AVLMs), such as GPT-4o~\cite{gpt4o}, have extended the capabilities of Vision-Language Models (VLMs) by incorporating the auditory modality.
This integration enables a holistic understanding of dynamic scenes involving both visual and acoustic cues~\cite{avqa}.
Existing approaches typically employ separate encoders for vision and audio, subsequently injecting tokens from both modalities into the LLM via concatenation~\cite{vl1,vl2}, cross-modal attention~\cite{me1}, or interleaving schemes~\cite{vs2,qwen3o}.
While these unified autoregressive frameworks~\cite{vl1, vs1, qwen25o} have significantly enhanced representation learning, they generally process audio and visual tokens within a single stream, implicitly assuming that dense interaction is always beneficial. This assumption, however, overlooks the potential risks of interference between modalities during the decoding process~\cite{avh}.

\subsection{Cross-Modal Interference and Hallucination} 
\label{sec:rel_hallucination} 
Despite the success of AVLMs, they are prone to cross-modal hallucinations, where the model generates content conditioned on one modality that contradicts or is unsupported by the other~\cite{avh}. 
This interference often stems from uncontrolled cross-modal interactions and modality bias~\cite{as1,as2,as3,inter1,inter2}, leading the model to hallucinate visual events from audio cues or vice versa.
Current mitigation strategies fall into two main categories: inference-time intervention and training-time alignment. 
Inference-time methods, such as AVCD~\cite{as1}, employ contrastive decoding to calibrate generation. Conversely, training-time approaches like OmniDPO~\cite{h3} utilize preference optimization to align models with audio-visual evidence.
However, these methods often rely on heuristic constraints or expensive annotated data, without fundamentally addressing the structural entanglement of audio-visual information that causes the interference~\cite{inter1,inter2}.
More recently, RL-CoMM~\cite{avconfuse} studies the confusion problem under artificially constructed asymmetric audio-visual inputs and employs an external audio teacher model to correct model confusion. In contrast, our work targets the more general cross-modal interference problem and proposes a solution from the perspective of internal reasoning process control. The two approaches differ substantially in problem definition, methodological core, and applicability scope.
\begin{figure*}[t]
  \centering
  \includegraphics[width=\textwidth]{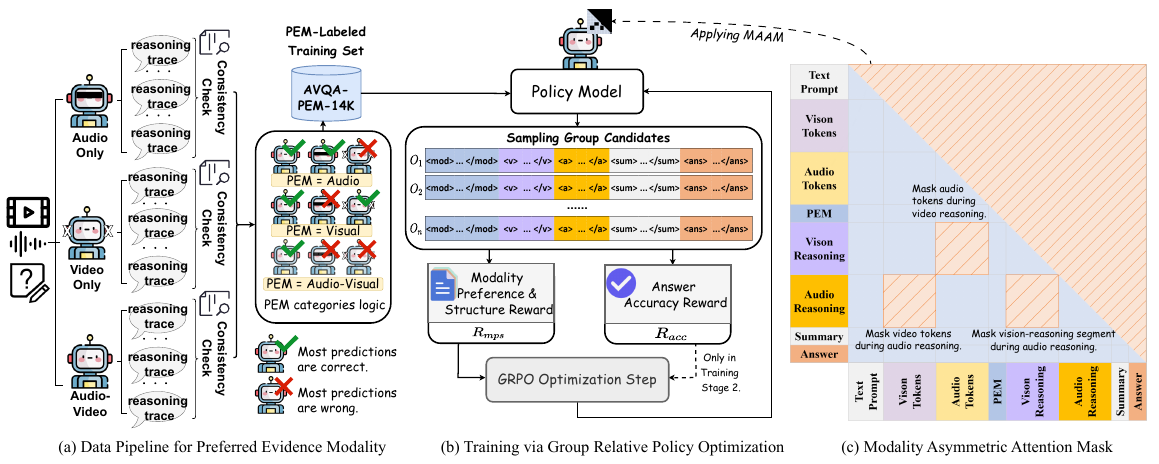}
  \caption{Overview of Separate First, Fuse Later (SFFL) reasoning framework.}
  \label{fig:SFFL}
\end{figure*}
\subsection{Multimodal Chain-of-Thought and Reasoning}
\label{sec:rel_cot} 
Chain-of-Thought (CoT) prompting has become a standard paradigm for enhancing the reasoning capabilities of Multimodal Large Language Models (MLLMs)~\cite{mcot,rl1,rl2,rl3,rl4}.
By explicitly generating intermediate reasoning traces, models can avoid exploiting prior-driven shortcuts when answering complex queries~\cite{short1,short2}.
Recent works have extended this into two-stage frameworks, such as Multimodal-CoT~\cite{mcot}, which generate a rationale before the final answer to improve robustness.
Furthermore, refinement and verification strategies like Vision-SR1~\cite{rl2} and MoT~\cite{rl3} attempt to validate reasoning paths or verify information across multiple perspectives to mitigate hallucinations.
Different from these approaches, our work explicitly isolates the information interaction of audio and visual throughout both training and inference. This structural separation prevents cross-modal shortcuts at the source, rather than merely correcting them post-hoc.

\section{Proposed Methods}
% In this section, we present SFFL in detail. Specifically, we first introduce the data pipeline for constructing Preferred Evidence Modality (PEM) in Section \ref{sec:pem}. Subsequently, in Section\ref{sec:savt} and Section\ref{sec:maam}, we respectively explain the Separate Thinking and Modality Asymmetric Attention Mask (MAAM) to enforce modality isolation during the chain-of-thought stage.
% Finally, we detail the reinforcement learning procedure used in training in Section \ref{sec:rl}. The method overview is shown in Fig.~\ref{fig:SFFL}.
% In this section, we present the Separate First, Fuse Later (SFFL) framework, as shown in Figure \ref{fig:SFFL}. To realize our Separate-then-Fuse reasoning paradigm, we construct training data with Preferred Evidence Modality (PEM) labels, which will be used in the subsequent Group Relative Policy Optimization (GRPO) training to guide evidence preference. For the model, we employ a Modality Asymmetric Attention Mask (MAAM) to ensure minimal cross-modal interference between audio and video modalities during the reasoning process.

In this section, we present the Separate First, Fuse Later (SFFL) framework, as shown in Figure~\ref{fig:SFFL}. To realize our Separate-then-Fuse reasoning paradigm, we construct training data with Preferred Evidence Modality (PEM) labels. Moreover, we employ a Modality Asymmetric Attention Mask (MAAM) to restrict cross-modal interaction, ensuring minimal audio–video interference during reasoning. We then perform a two-stage Group Relative Policy Optimization (GRPO) procedure with two reward components: Modality Preference \& Structure reward, and Answer Accuracy reward.

\subsection{Data Pipeline for Preferred Evidence Modality}
\label{sec:pem} 
% Existing AVLMs are broadly vision-dominant during multimodal reasoning~\cite{as1,as2,bias1}. However, the evidential contribution of different modalities varies significantly across samples, creating a need for instance-specific modality preference.~\cite{avqam2,avqam4}. We seek to enable the model to identify its modality evidence preference during reasoning, referred to as the Preferred Evidence Modality (PEM). It serves to characterize the model’s intrinsic reasoning preference while preserving full multimodal inputs.
% \textbf{TODO We adopt self-distillation because PEM is both model-specific and instance-specific.
% Specifically, the use of modality cues in a model is shaped by its inductive biases and can vary across instances, making it hard for hand-crafted rules or external annotations to align with its actual reasoning behavior~\cite{self1,self2}.}

To mitigate the dominance of visual information in AVLMs~\cite{as1,as2,bias1} and enable the model to adaptively identify and rely on the modality that offers the most decisive evidence during reasoning, we introduce the Preferred Evidence Modality (PEM). PEM helps the model identify its modality preference while preserving full multimodal inputs.

We build an AVQA-based annotation pipeline~\cite{avqa} (Fig.~\ref{fig:SFFL}) to automatically obtain about 14k high-confidence PEM-labeled instances for subsequent training.
For each original AVQA instance (a question and its corresponding audio-video clip), we construct three modality input settings:
Audio-only (A), Video-only (V), and Audio-Visual (AV).
For each setting, we use Qwen3-Omni-Instruct~\cite{qwen3o} to generate chain-of-thought (CoT) answers with multiple sampling runs, yielding $n$ candidate answers $\{\hat{y}_k\}_{k=1}^{n}$ and their CoT traces $\{r_k\}_{k=1}^{n}$.
We deem the question can be solved under the current input setting when:
(i) the correctness rate over the $n$ samples is at least $\tau_{\mathrm{acc}}$,
\begin{equation}
    \frac{1}{n}\sum_{k=1}^{n}\mathbf{1}\!\left[\hat{y}_k = y\right] \ge \tau_{\mathrm{acc}},
\end{equation}
and (ii) the CoT consistency across the $n$ samples is at least $\tau_{\mathrm{cons}}$, measured as the average embedding similarity over all pairwise comparisons between CoT texts~\cite{con}.
We label each instance into one of three PEM categories:
\begin{itemize}
  \item PEM = Audio: If both A and AV can solve the question, while V cannot, we label it as Audio.
  \item PEM = Visual: If both V and AV can solve the question, while A cannot, we label it as Visual.
  \item PEM = Audio-Visual: If neither A nor V can solve the question, but AV can, we label it as Audio-Visual.
\end{itemize}

We discard ambiguous, contradictory, or trivially solvable cases (e.g., instances solvable under all input settings), retaining only instances with consistent solvability patterns for labeling. We use this controlled modality probing only for annotation, while subsequent training is conducted with full multimodal inputs.
% We store PEM as the reward signal for subsequent reinforcement learning. In practice, the generated chain-of-thought (CoT) is used only for later SFT comparison experiments.
\subsection{Separate-then-Fuse AV Reasoning}
\label{sec:sfr} 

Inspired by the new parallel thinking capabilities of Google Gemini~\cite{gemini} and the insights from Parallel-R1~\cite{p1} in exploring and learning parallel thinking behaviors in general mathematical reasoning tasks, we aim to explicitly separate the reasoning processes of audio and video in a structured chain-of-thought to reduce cross-modal interference. Therefore, we propose Separate-then-Fuse AV Reasoning (SFR).

SFR is implemented with lightweight control tags that explicitly structure the reasoning process,
including \texttt{<mod>, <v>, <a>, <sum>, <ans>} and their corresponding closing tags
\texttt{</mod>, </v>, </a>, </sum>, </ans>}.
Following the structured output formulation in Fig~\ref{fig:intro}, we augment the model's textual output with these boundary tags to mark regions that require PEM decision, modality-isolated reasoning, and fusion.
Specifically, the model output is organized as
\[
\begin{aligned}
&[\texttt{<mod>},\, m,\, \texttt{</mod>}, \\
&\;\texttt{<v>},\, v_0,\ldots,v_p,\, \texttt{</v>},\;\texttt{<a>},\, a_0,\ldots,a_q,\, \texttt{</a>}, \\
&\;\texttt{<sum>},\, s_0,\ldots,s_r,\, \texttt{</sum>},\; \texttt{<ans>},\, y,\, \texttt{</ans>}]
\end{aligned}
\]
where $m$ denotes the Preferred Evidence Modality (PEM) (Section~\ref{sec:pem}),
$\{v_t\}_{t=0}^{p}$ and $\{a_t\}_{t=0}^{q}$ denote the visual- and audio-grounded reasoning tokens, respectively,
$\{s_t\}_{t=0}^{r}$ denotes a concise fused summary, and $y$ denotes the final prediction.
The control tags \texttt{<v>} and \texttt{<a>} are used to enforce strictly separated reasoning traces, preventing evidence from one modality from biasing the interpretation of the other,
while \texttt{<mod>} provides an instance-dependent preference signal that guides the subsequent fusion.
Finally, \texttt{<sum>} denotes the fusion process, followed by \texttt{<ans>} for answer prediction.

\subsection{Modality Asymmetric Attention Mask}
\label{sec:maam}

Although SFR explicitly delineates modality-specific reasoning regions via control tags, the underlying causal attention mechanism poses a subtle challenge: by default, every token can attend to all preceding tokens, which introduces undesirable information leakage.
When generating the visual reasoning segment \texttt{<v>}$,v_0,\ldots,v_p,$\texttt{</v>}, the model may still attend to audio input tokens $A$, contaminating the intended visual-only reasoning.
Likewise, when generating the audio reasoning segment \texttt{<a>}$,a_0,\ldots,a_q,$\texttt{</a>}, the model may attend to video input tokens $V$ or reuse information from the previously generated visual reasoning segment, resulting in cross-modal information leakage.
To mitigate this, we propose the Modality Asymmetric Attention Mask (MAAM), which imposes asymmetric visibility constraints on different reasoning segments, explicitly controlling the interaction of information, as illustrated in Fig~\ref{fig:SFFL}.

\subsubsection{Formal Definition}

Let the total sequence length be $L$. The attention mask $\mathbf{M} \in \mathbb{R}^{L \times L}$ controls which key positions (column $j$) each query position (row $i$) can attend to.
The standard causal attention mask is defined as
\begin{equation}
\mathbf{M}_{ij}^{\mathrm{causal}} =
\begin{cases}
0, & j \le i,\\
-\infty, & j > i.
\end{cases}
\end{equation}
MAAM augments this base mask with an additional modality-aware mask $\mathbf{M}^{\mathrm{MAAM}}$, yielding the composite attention mask
\begin{equation}
\mathbf{M}_{ij} = \mathbf{M}_{ij}^{\mathrm{causal}} + \mathbf{M}_{ij}^{\mathrm{MAAM}}.
\end{equation}

To specify the masking rules, we define the following token position sets: 
$K^V$ for video input tokens $V$, $K^A$ for audio input tokens $A$, $Q^v$ for visual reasoning tokens $\{v_t\}_{t=0}^{p}$ inside \texttt{<v>}$\ldots$\texttt{</v>}, $Q^a$ for audio reasoning tokens $\{a_t\}_{t=0}^{q}$ inside \texttt{<a>}$\ldots$\texttt{</a>}, and $K^v$ for all positions within the \texttt{<v>}$\ldots$\texttt{</v>} span (including $Q^v$ and the boundary tags).

% \begin{itemize}
%     \item $K^V$: positions of video input tokens $V$,
%     \item $K^A$: positions of audio input tokens $A$,
%     \item $Q^v$: positions of visual reasoning tokens $\{v_t\}_{t=0}^{p}$ inside \texttt{<v>}$\ldots$\texttt{</v>},
%     \item $Q^a$: positions of audio reasoning tokens $\{a_t\}_{t=0}^{q}$ inside \texttt{<a>}$\ldots$\texttt{</a>},
%     \item $K^v$: all positions within the \texttt{<v>}$\ldots$\texttt{</v>} span, including $Q^v$ and boundary tags.
% \end{itemize}

The core principle of MAAM is to enforce asymmetric visibility by selectively blocking certain query–key pairs, as illustrated in Fig~\ref{fig:SFFL}.
When visual reasoning tokens $\{v_t\}_{t=0}^{p}$ serve as queries, they are prevented from attending to audio input tokens $A$.
When audio reasoning tokens $\{a_t\}_{t=0}^{q}$ serve as queries, they are prevented from attending to both video input tokens $V$ and the entire visual reasoning span, thereby avoiding cross-span leakage.
Formally,
\begin{align}
\mathbf{M}_{ij}^{\mathrm{MAAM}} &= -\infty, \quad \forall\, i \in Q^v,\; \forall\, j \in K^A, \\
\mathbf{M}_{ij}^{\mathrm{MAAM}} &= -\infty, \quad \forall\, i \in Q^a,\; \forall\, j \in K^V, \\
\mathbf{M}_{ij}^{\mathrm{MAAM}} &= -\infty, \quad \forall\, i \in Q^a,\; \forall\, j \in K^v.
\end{align}
Beyond these constraints, all other query–key pairs follow the default causal visibility, i.e., $\mathbf{M}_{ij}^{\mathrm{MAAM}} = 0$.
% \subsubsection{Mask Construction}
We apply MAAM by scanning the sequence to identify token boundaries.
Specifically, we locate $K^V$ and $K^A$ using modality indicator tokens, and identify $Q^v$, $Q^a$, and $K^v$ via the control tags.
We then instantiate a boolean attention matrix and fuse it with the causal mask.
During training, an $L \times L$ mask is constructed once per sample, which is then broadcast across attention heads.
During autoregressive inference, MAAM does not construct the full $L \times L$ matrix. At decoding step $t$, only the mask row for the current query token is updated, using cached boolean indicators that record the token type of each preceding position. Based on the current token type (e.g., visual reasoning, audio reasoning, or fusion), simple element-wise masking rules are applied to determine which key positions are visible. Consequently, the per-step additional computation acts only on a vector of length $t$, yielding a per-step overhead of $O(t)$. In the worst case, the total overhead across all $L$ decoding steps sums to $O(L^2)$, but the constant factor is extremely small since it involves only boolean operations rather than dense matrix computation.

% for all query positions $i \in Q^v$, we mask the key positions $K^A$;
% for all query positions $i \in Q^a$, we mask the key positions $K^V \cup K^v$.

\subsection{Reward for Reinforcement Learning}
\label{sec:rl} 
% DeepSeek-R1-Zero~\cite{rl0} demonstrates that reinforcement learning can robustly induce reasoning behaviors when the training signal is grounded in verifiable reward.
% This motivates our design: instead of using SFT to impose the desired reasoning pattern, we construct two verifiable reward functions that directly optimize for the Separate-then-Fuse reasoning process described in Section~\ref{sec:sfr}.
DeepSeek-R1-Zero~\cite{rl0} shows that reinforcement learning with verifiable rewards can robustly induce reasoning behaviors. Following this, instead of SFT, we construct two verifiable reward functions that directly optimize the Separate-then-Fuse reasoning process (Section~\ref{sec:sfr}).

\textbf{Modality Preference \& Structure Reward.} Verifies that the model correctly identifies the Preferred Evidence Modality (PEM) and follows the Separate-then-Fuse reasoning structure:
\begin{equation}
R_{\mathrm{mps}} =
\begin{cases}
1, & \text{if PEM is correct \& structure matches},\\
0, & \text{otherwise}.
\end{cases}
\end{equation}

\textbf{Answer Accuracy.} Checks whether the model's answer matches the ground truth:
\begin{equation}
R_{\mathrm{acc}} =
\begin{cases}
1, & \text{if Answer = GT},\\
0, & \text{otherwise}.
\end{cases}
\end{equation}

We adopt a two-stage Group Relative Policy Optimization (GRPO) scheme: Stage~1 optimizes $R_{\mathrm{mps}}$, and Stage~2 optimizes a weighted sum of $R_{\mathrm{acc}}$ and $R_{\mathrm{mps}}$.

\begin{equation}
R_{stage1} = R_{\mathrm{mps}} \, .
\end{equation}
\begin{equation}
R_{stage2} = \lambda_{\mathrm{acc}} \cdot R_{\mathrm{acc}} + \lambda_{\mathrm{mps}} \cdot R_{\mathrm{mps}} \, .
\end{equation}

% \clearpage

\begin{table*}[!t]
  \centering
    \begin{small}
      \begin{sc}
        \setlength{\tabcolsep}{2.5pt}
        \begin{tabular}{lcccccccc}
          \toprule
          \multirow{2}{*}{Methods} &
          \multicolumn{4}{c}{AVHBench} &
          \multicolumn{4}{c}{General AVQA} \\
          \cmidrule(lr){2-5}\cmidrule(lr){6-9}
          & VAH$\uparrow$ & AVH$\uparrow$ & MIS$\uparrow$ & Avg.$\uparrow$ & AVQA$\uparrow$ & Valor2$\uparrow$ & \makecell{\small MUSIC\\-AVQA$\uparrow$} & Avg.$\uparrow$ \\
          
          \midrule
          Qwen3-Omni-Thinking~\cite{qwen3o}
          & 77.12 & 82.31 & 70.68 & 75.95
          & 90.79 & 75.59 & 63.64 & 76.49 \\
          VideoLLaMA2.1-7B-AV~\cite{vl2}
          & 73.93 & 61.71 & 51.76 & 63.47
          & 85.69 & 60.57 & 79.32 & 68.65 \\
          video-SALMONN-2+(7B)~\cite{vs2}
          & 56.27 & 84.26 & 49.68 & 59.94
          & 80.69 & 61.43 & 63.41 & 66.37 \\
          Gemini-3-Flash~\cite{gemini}
          & 72.14 & 71.65 & 72.81 & 72.27
          & 89.34 & 72.67 & 58.81 & 73.27 \\
          RL-CoMM~\cite{avconfuse}$^{\dagger}$
          & 65.63 & 78.96 & 51.85 & 65.48
          & \textbf{95.87} & - & 79.46 & - \\
          VideoLLaMA2-AVCD~\cite{as1}$^{\dagger}$
          & - & - & - & 72.15
          & - & - & \textbf{81.58} & - \\
          \addlinespace[0.5em]
          \multicolumn{8}{l}{\textbf{Backbone model: Qwen2.5-Omni-7B~\cite{qwen25o}}} \\
          \addlinespace[0.2em]
          Zero-shot Inference
          & 61.41 & 70.02 & 61.51 & 63.29
          & 88.07 & 66.36 & 58.82 & 69.14 \\
          SFFL (our)
          & 62.27 & 78.61 & 59.49 & 64.79
          & 88.67 & 70.59 & 62.71 & 71.69\\
          \addlinespace[0.5em]
          \multicolumn{8}{l}{\textbf{Backbone model: Qwen3-Omni-30B-A3B-Instruct~\cite{qwen3o}}} \\
          \addlinespace[0.2em]
          Zero-shot Inference
          & 74.28 & 81.95 & 66.36 & 73.12
          & 89.62 & 76.56 & 66.00 & 76.33\\
          AVQA-PEM-14K data (GRPO)
          & 75.20 & 81.69 & 73.08 & 75.84
          & 91.31 & 76.35 & 66.61 & 77.53\\
          SFFL (our)
          & \textbf{80.79} & \textbf{85.12} & \textbf{79.58} & \textbf{81.29}
          & 92.31 & \textbf{77.43} & 69.93  & \textbf{80.24} \\
          \bottomrule
        \end{tabular}
      \end{sc}
    \end{small}
  \caption{Comparison of Our Method (SFFL) with Baselines on AVHBench and General AVQA Benchmarks. Results marked with a superscript $^{\dagger}$ are taken from the original papers}
  \label{tab:main}
\end{table*}

\section{Experiment}
% \subsection{Baseline Methods}
% We select Qwen3-Omni-Thinking, Qwen3/2.5-Omni-Instruct~~\cite{qwen3o,qwen25o}, VideoLLaMA2.1-AV~~\cite{vl2}, and video-SALMONN-2+~~\cite{vs2} as our baseline models, covering mainstream open-source audio-visual LLMs. We further implement and evaluate our method on Qwen3\allowbreak/\allowbreak Qwen2.5-Omni-Instruct to achieve a practical balance between performance and reproducibility.

\subsection{Dataset}
We use the AVQA-PEM-14K dataset constructed in Section~\ref{sec:pem} as our training set. Our evaluation measures the AVQA task capabilities of Audio-Visual Language Models (AVLMs) from two perspectives: cross-modal hallucination, and general audio-visual QA.
% \subsubsection{Cross-modal hallucination}
We use AVHBench~\cite{avh} to assess cross-modal hallucination and we only retain its QA tasks for evaluation. AVHBench provides around 5K QA pairs to evaluate a model’s ability to identify hallucinated objects and events in given audio-visual inputs, or to judge whether the audio and visual signals correspond to each other.
It includes three task types: Video-driven Audio Hallucination, Audio-driven Video Hallucination, and Audio-visual Matching (VAH, AVH, MIS).
% \subsubsection{General Audio-Visual QA}
We evaluate general audio-visual QA on three diverse benchmarks. 
AVQA~\cite{avqa} is a large-scale Audio-Visual QA benchmark targeting joint audio-visual reasoning in real-world scenarios. AVQA-PEM-14K is built on AVQA, so the AVQA test set is mainly used to measure in-domain performance.
Valor32k-AVQA v2.0 (Valor2)~\cite{valor2} contains around 30K real-world videos and over 200K QA pairs, and we adopt it as a more challenging complementary benchmark to test generalization and robustness.
MUSIC-AVQA~\cite{musicavqa} is a domain-specific audio-visual question answering benchmark built on real-world music performance videos. We include it to assess the model’s open-ended audio-visual reasoning capability.
% AVUT focuses on audio-driven video understanding, and mitigates shortcut issues through an answer permutation-based filtering mechanism, thereby more strictly evaluating genuine comprehension of audio content and audio-visual interactions.

\subsection{Training Details}
Our primary experiments are conducted with Qwen3-Omni-30B-A3B-Instruct~\cite{qwen3o}. Our codebase is adapted from ms-swift~\cite{swift}, and we largely follow its official GRPO training recipe. Training is conducted in two stages. 
In Stage 1, we run GRPO with LoRA on our curated AVQA-PEM-14K, using a global batch size of 96, a learning rate of 1e-5, 4 rollouts, and a Modality Preference \& Structure reward. In Stage 2, we continue training on the same dataset with a batch size of 48, a learning rate of 1e-6, and 4 rollouts, where the reward is a weighted combination of Modality Preference \& Structure reward and Answer Accuracy reward with weights 0.2 and 1.0, respectively. Our training is performed on 8 NVIDIA H20 GPUs.

\subsection{Main Results}
Table~\ref{tab:main} reports results on AVHBench (VAH, AVH, MIS) and three general AVQA benchmarks (AVQA, Valor32k-AVQA v2.0, MUSIC-AVQA), all by answer accuracy. We compare against mainstream audio-visual LLMs (Qwen3-Omni-30B-A3B-Thinking~\cite{qwen3o}, VideoLLaMA2.1-7B-AV~\cite{vl2}, video-SALMONN-2+~\cite{vs2}, Gemini-3-Flash~\cite{gemini}), the contrastive decoding (CD) framework~\cite{as1}, and a GRPO RL baseline with only Answer Accuracy Reward. For generalization, we report both zero-shot and our method on two backbones, Qwen2.5-Omni-7B~\cite{qwen25o} and Qwen3-Omni-30B-A3B-Instruct~\cite{qwen3o}.

Overall, SFFL delivers consistent improvements across both backbones. On Qwen3-Omni-30B-A3B-Instruct, it achieves an average AVHBench score of 81.29 and scores of 92.31, 77.43, and 69.93 on AVQA, Valor2, and MUSIC-AVQA, respectively, yielding an overall average of 80.24. Its relatively weaker MUSIC-AVQA result may reflect differences in the training-data bias of the VideoLLaMA2 family. On Qwen2.5-Omni-7B, SFFL also achieves a general AVQA average of 71.69, demonstrating consistent gains at the 7B scale.
Compared with the closely related RL-CoMM~\cite{avconfuse}, SFFL achieves higher scores across all AVHBench subsets, with the largest gain on MIS (+27.73), demonstrating the benefit of explicit reasoning-time information-flow control. As the backbones and training settings differ, the comparison is for reference.
These results demonstrate that SFFL consistently outperforms most prior baselines on both hallucination metrics and general audio-visual QA reasoning tasks. By employing a modality-specific reasoning paradigm, it effectively mitigates the issues of cross-modal interference, thereby enhancing the model's inherent reasoning generalization performance.

\subsection{Validating Preferred Evidence Modality}
\subsubsection{Train-Free Controlled Experiment}
\begin{table}[t]
  \begin{center}
    \begin{small}
      \begin{sc}
        \setlength{\tabcolsep}{2pt}
        \renewcommand{\arraystretch}{0.95}
        \begin{tabular}{c p{4.5cm}ccc}
          \toprule
          ID & Setting & VAH$\uparrow$ & AVH$\uparrow$ & MIS$\uparrow$ \\
          \midrule
          1 & Audio-only input
            & \textbf{80.09} & -- & 50.64 \\
          2 & Video-only input
            & -- & \textbf{83.71} & 48.24 \\
          3 & AV input
            & 74.28 & 81.95 & 66.36 \\
          4 & AV input, PEM=Audio
            & 79.43 & 81.34 & 71.91 \\
          5 & AV input, PEM=Visual
            & 75.46 & 83.27 & \textbf{73.13} \\
          6 & AV input, PEM=Audio-Visual
            & 75.46 & 81.69 & 71.86 \\
          \bottomrule
        \end{tabular}
      \end{sc}
    \end{small}
  \end{center}
  \caption{Train-free controlled experiment on AVHBench: we evaluate unimodal (A-only/V-only), multimodal (AV), and AV with forced PEM (Audio, Visual, Audio-Visual) settings on VAH (audio-dominant), AVH (visual-dominant), and MIS (audio-visual matching, requiring consistency between audio and video evidence).}
  \label{tab:ablation_rat}
\end{table}
% We first conduct a train-free controlled experiment to validate the rationale of Preferred Evidence Modality (PEM): for some instances, a particular modality provides the primary evidence required for answering. Accordingly, we expect the PEM to guide the model to prefer that modality evidence during reasoning.
% We choose AVHBench as the evaluation set because its three task types are naturally aligned with our PEM motivation. In particular, Video-driven Audio Hallucination (VAH) and Audio-driven Video Hallucination (AVH) are constructed so that answering the question mainly depends on the corresponding dominant modality evidence (audio evidence for VAH and visual evidence for AVH), whereas Audio-Visual Matching (MIS) requires jointly reasoning over both audio and visual evidence to judge whether the audio and visual information are consistent with each other.
% --- 压缩 (保VAH/AVH/MIS定义与证据方向, 收紧冗余表述) ---
We first conduct a train-free controlled experiment to validate the rationale of Preferred Evidence Modality (PEM): for some instances, a particular modality provides the primary evidence, and we expect PEM to guide the model toward it.
We choose AVHBench because its three task types align with our PEM motivation: Video-driven Audio Hallucination (VAH) and Audio-driven Video Hallucination (AVH) mainly depend on a single dominant modality (audio for VAH, visual for AVH), while Audio-Visual Matching (MIS) requires joint audio-visual reasoning to judge cross-modal consistency.

Specifically, without any training, we evaluate the same backbone under six inference settings: (1) Audio-only input, (2) Video-only input, (3) Audio with Video (AV) input, and (4-6) AV input with the PEM forcibly set to Audio, Visual, or Audio-Visual, respectively.
As shown in Table~\ref{tab:ablation_rat}, Setting (1) (A-only) and Setting (2) (V-only) achieve relatively high unimodal accuracy. This indicates that, for these questions, a particular modality often provides the primary evidence needed for a correct answer.
However, directly feeding Setting (3) (AV input) reduces the overall performance, suggesting noticeable cross-modal interference when an additional modality is introduced. More importantly, with the AV input kept unchanged, forcing PEM to Audio (Setting (4)) or Visual (Setting (5)) raises VAH/AVH from 74.28 to 79.43 and from 81.95 to 83.27, respectively. This suggests that PEM helps suppress cross-modal interference. In addition, Setting (5) attains the highest MIS accuracy, while Setting (6) does not yield the best MIS, which further supports our motivation: many AVLMs exhibit a vision-dominant bias, and naïvely combining audio and video reasoning may cause cross-modal interference.

% In contrast, forcing \texttt{<mod>} to AV further degrades Acc.~AV to 62.26, confirming that unnecessary multimodal fusion can hurt reasoning performance.
\subsubsection{Effectiveness of AVQA-PEM-14K and PEM Training}
\begin{table}[t]
  \centering
  \begin{small}
    \begin{sc}
      \setlength{\tabcolsep}{2pt}
      \renewcommand{\arraystretch}{0.95}
      \begin{tabular}{p{2.7cm}cccc}
        \toprule
        Method & PEM Acc.$\uparrow$ & VAH$\uparrow$ & AVH$\uparrow$ & MIS$\uparrow$ \\
        \midrule
        Origin w/ CoT
         & 87.76 & 75.41 & 79.93 & 74.89 \\
        \addlinespace[0.4em]
        \multicolumn{5}{l}{\textbf{Ours Stage 1}} \\
        Random
         & 33.3 & 73.41 & 74.21 & 64.16 \\
        Oracle-PEM
         & 100.00 & \textbf{79.69} & \textbf{84.68} & \textbf{78.94} \\
        Predicted-PEM
         & 94.40 & 76.64 & 81.78 & 78.25 \\
        \bottomrule
      \end{tabular}
    \end{sc}
  \end{small}
  \caption{Effectiveness of AVQA-PEM-14K PEM training. We report PEM accuracy and task performance (VAH/AVH/MIS) under different PEM sources at inference time (Random, Oracle, Predicted).}
  \label{tab:PEM_eff}
\end{table}

To validate the effectiveness of our constructed AVQA-PEM-14K dataset and the corresponding PEM training stage, we further conduct an ablation study. Specifically, we keep the Separate-then-Fuse Reasoning prompting template and systematically compare different PEM settings to quantify their impact on performance. Results are reported in Table~\ref{tab:PEM_eff}.
First, compared with the original backbone, the model after the Stage 1 training (using Modality Preference \& Structure Reward) achieves consistent improvements in both PEM accuracy and task accuracy.
This indicates that training on AVQA-PEM-14K effectively strengthens the model’s ability to identify modality evidence requirements and yields tangible gains on the task.
Next, to isolate the effect of PEM selection from other factors, we fix the Stage 1 checkpoint and vary only how PEM is obtained at inference time. We compare three alternatives: Random-PEM injected directly via prompts, Oracle-PEM using ground-truth labels, and Predicted-PEM generated by the model. 
Oracle-PEM serves as an upper bound, characterizing the attainable performance of Separate-then-Fuse Reasoning under perfect modality evidence requirement identification.
Predicted-PEM closely approaches this upper bound, demonstrating that the modality recognition capability learned from AVQA-PEM-14K is sufficient to support Separate-then-Fuse Reasoning and achieve near-optimal reasoning performance.
In contrast, Random-PEM performs substantially worse, further indicating that appropriate PEM selection has an impact on reasoning.

\subsubsection{Interpretability of Preferred Evidence Modality}
\begin{figure}[t]
  \centering
  \includegraphics[width=\columnwidth]{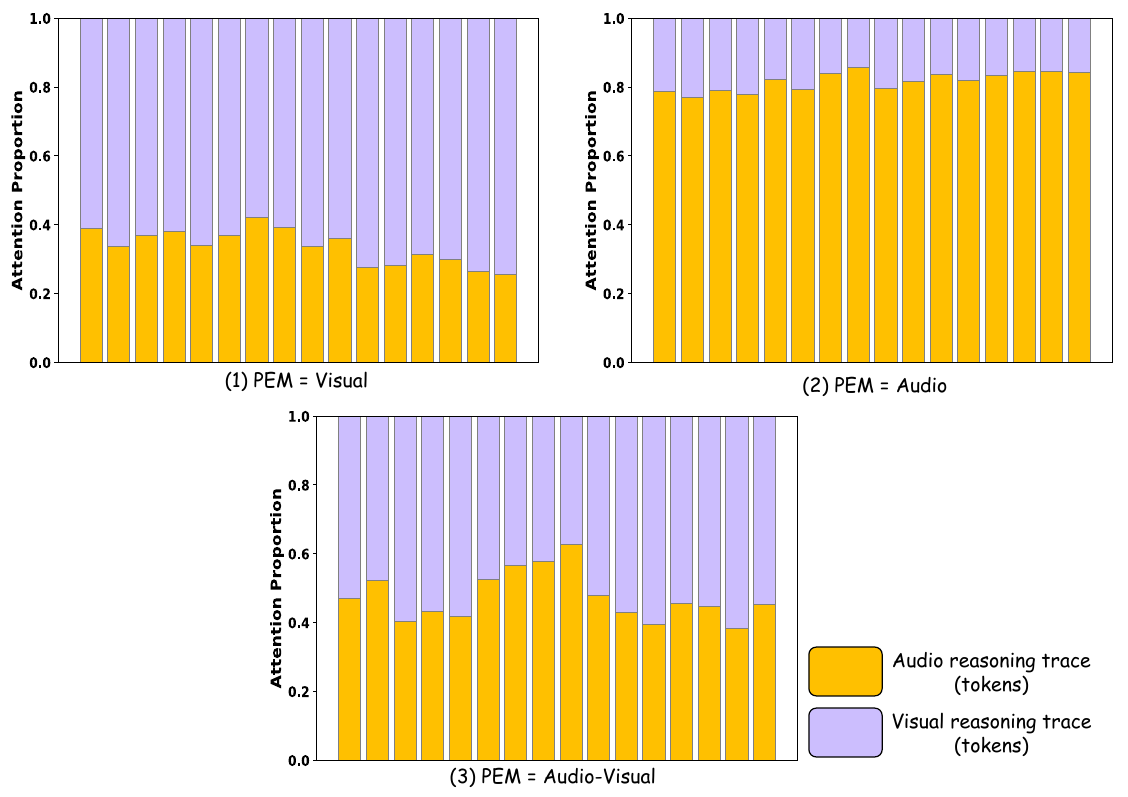}
  \caption{Layer-wise attention allocation from \texttt{<sum>} tokens to audio vs. visual reasoning traces. We report the normalized attention mass across the last 16 layers, grouped by predicted PEM.}
  \label{fig:inter}
\end{figure}
We further validate the interpretability of PEM and examine how it influences the model’s reasoning.
Specifically, as shown in Fig~\ref{fig:inter}, we measure the attention allocation during the generation of the final integrated text (\texttt{<sum>}), i.e., how the model attends to the audio versus visual reasoning traces (\texttt{<v></v>},\texttt{<a></a>}). We report the normalized proportions over the last 16 layers and group samples by the model-predicted PEM. The results show a clear alignment: when PEM indicates Audio, attention during integration concentrates more on the audio-related reasoning span; when PEM indicates Visual, attention shifts markedly toward the visual-related span; and when PEM is Audio-Visual, the attention allocation becomes more balanced across modalities. This consistency suggests that PEM is not an arbitrary label, but instead reflects the model’s instance-level evidence reliance and integration pathway, providing interpretable support for using PEM as a training signal.

Taken together, the above findings suggest that the backbone already has the capacity to answer correctly when relying on the appropriate unimodal evidence, but its performance is hindered by cross-modal interference under naïve audio-visual inference. Therefore, our subsequent SFFL training does not aim to inject new knowledge. Instead, it aims to align the model’s reasoning with the PEM-guided evidence path, thereby unlocking this latent capability.
% Overall, these results validate the core motivation of PEM: mitigating the prevalent vision-dominant bias in multimodal reasoning, thereby improving the reliability and generalization of audio-visual reasoning.
\subsection{Ablation Study}
\subsubsection{Ablation on Separate-then-Fuse AV Reasoning and Modality Asymmetric Attention Mask}

% TODO caption
% TODO format
% TODO important word
To verify the contribution of each core design to the performance gains, we conduct ablation studies on AVHBench and AVQA. We consider two model variants: a zero-shot baseline without any training, and a GRPO-trained variant optimized with GRPO on AVQA-PEM-14K. Within each variant, we ablate our two key components, Separate-then-Fuse AV Reasoning (SFR) and the Modality Asymmetric Attention Mask (MAAM), to isolate their individual effects.

% \textbf{Separate-then-Fuse AV Reasoning:}
In the GRPO-trained variant, enabling SFR alone improves MIS from 73.08 to 78.50 (+5.42), while the other metrics show only marginal gains. This indicates that SFR primarily mitigates errors caused by cross-modal interference, which is most directly reflected by MIS.
In the zero-shot variant, SFR yields an even larger improvement on MIS (+8.53), but causes an unstable drop in AVH (-2.02) and almost no change in AVQA (-0.01). This suggests that CoT-style separation alone is insufficient to fully prevent cross-modal leakage: erroneous evidence may still merge during generation and mislead the final answer, leading to fluctuations on certain tasks.

% \textbf{Modality Asymmetric Attention Mask:}
Introducing MAAM brings consistent and significant gains on AVH and AVQA. For example, in the zero-shot variant, AVQA improves from 89.61 to 92.11 (+2.50) and AVH from 79.93 to 83.98 (+4.05). These results show that MAAM’s explicit blocking mechanism effectively suppresses cross-modal interference, improving robustness and stabilizing the potential benefits of SFR.

Overall, MAAM is the main source of stable improvements, whereas SFR mainly boosts MIS by reducing interference. Training on AVQA-PEM-14K further improves stability and the upper-bound performance on hallucination-related metrics and AVQA tasks.

\subsubsection{Ablation on Training Approach}
\begin{table}[t]
  \centering
  \begin{small}
    \begin{sc}
      \setlength{\tabcolsep}{0.5pt}
      \renewcommand{\arraystretch}{0.95}
      \begin{tabular}{p{1.55cm}ccccc c}
        \toprule
        \multirow{2}{*}{Settings} &
        \multirow{2}{*}{w/SFR} &
        \multirow{2}{*}{w/MAAM} &
        \multicolumn{3}{c}{AVHBench} &
        \multirow{2}{*}{AVQA$\uparrow$} \\
        \cmidrule(lr){4-6}
        & & & VAH$\uparrow$ & AVH$\uparrow$ & MIS$\uparrow$ & \\
        \midrule

        \multirow{3}{*}{\shortstack[l]{Train\\w/\\GRPO}}
          & \cmark & \cmark & \textbf{80.79} & \textbf{85.12} & \textbf{79.58} & \textbf{92.31} \\
          & \cmark & \xmark & 76.70 & 81.92 & 78.50 & 91.52 \\
          & \xmark & \xmark & 75.2  & 81.69 & 73.08 & 91.31 \\
        \midrule

        \multirow{3}{*}{\shortstack[l]{Train\\Free}}
          & \cmark & \cmark & 75.75 & 83.98 & 79.30 & 92.11 \\
          & \cmark & \xmark & 75.41 & 79.93 & 74.89 & 89.61 \\
          & \xmark & \xmark & 74.28 & 81.95 & 66.36 & 89.62 \\
        \bottomrule
      \end{tabular}
    \end{sc}
  \end{small}
  \caption{Ablation of Separate-then-Fuse AV Reasoning and the Modality Asymmetric Attention Mask for a train-free baseline and a GRPO-trained variant.}
  \label{tab:ablation_dt_maam}
\end{table}
\begin{table}[t]
  \centering
  \begin{small}
    \begin{sc}
      \setlength{\tabcolsep}{2pt}
      \renewcommand{\arraystretch}{0.95}
      \begin{tabular}{p{2.9cm}ccc}
        \toprule
        Setting & AVHBench$\uparrow$ & AVQA$\uparrow$ & Valor2$\uparrow$ \\
        \midrule
        \textit{Zero-shot}
        & 73.12 & 89.62 & 76.56 \\
        \textit{SFT Approach}
        & 74.80 & 82.94 & 68.54 \\
        \midrule
        \textit{GRPO Approach}
        & & & \\
        \hspace*{0.6em}Reward (ACC)
        & 75.84 & \textbf{92.62} & 73.37 \\
        \hspace*{0.6em}Reward (MPS)
        & 78.25 & 91.52 & 76.79 \\
        \hspace*{0.6em}Ours
        & \textbf{81.29} & 92.31 & \textbf{77.43} \\
        \bottomrule
      \end{tabular}
    \end{sc}
  \end{small}
  \caption{Ablation study on training strategy and reward.}
  \label{tab:ablation_reward}
\end{table}
To verify the contribution of our reward design and training strategy, we conduct ablation studies on AVHBench, AVQA, and Valor32k-AVQA v2.0 (Table~\ref{tab:ablation_reward}). All trained variants use AVQA-PEM-14K, and we compare an SFT baseline, GRPO with Answer Accuracy (ACC) only, GRPO with Modality Preference \& Structure (MPS) only, and our two-stage composite reward, with the zero-shot baseline for reference.

While SFT slightly improves AVHBench over the zero-shot baseline, it degrades AVQA and Valor32k-AVQA v2.0, as SFT on limited data overwrites the generalizable behaviors of the already-strong base model. Using ACC as the sole reward substantially improves AVQA but brings little gain on Valor32k-AVQA v2.0, indicating limited cross-dataset benefit from answer-only alignment. Replacing it with MPS yields more consistent improvements across all three benchmarks, as the PEM-guided reward mitigates modality dominance and improves reasoning reliability. Our two-stage composite reward achieves the best results overall, confirming the complementarity of the two objectives: Stage~1 aligns evidence preference via MPS, and Stage~2 adds ACC under this constraint to further improve answer correctness, balancing robustness and QA performance.
\section{Conclusion}
In this paper, we present the Separate First, Fuse Later (SFFL) framework to mitigate cross-modal interference in AVLMs for audio-visual question answering.
We construct training data with Preferred Evidence Modality labels to encourage instance-dependent modality preference when answering. Through two-stage Group Relative Policy Optimization, together with Separate-then-Fuse CoT prompting and a Modality Asymmetric Attention Mask, we realize a structured reasoning paradigm that enforces modality-isolated chain-of-thought generation and permits controlled evidence fusion only at the final stage. 
Experiments on hallucination-focused and general AVQA benchmarks consistently improve both accuracy and robustness, indicating an effective solution to cross-modal interference and the resulting hallucinations.

\newpage

\bibliography{example_paper}

\end{document}